\documentclass[runningheads]{llncs}
\usepackage{times}
\usepackage{epsfig}
\usepackage{graphicx}
\usepackage{amsmath}
\usepackage{amssymb}
\usepackage{balance}
\usepackage{epsfig}
\usepackage{caption}
\usepackage{subcaption}
\usepackage{booktabs}
\usepackage{xcolor, soul}
\sethlcolor{cyan}

\begin{document}
\title{Multimodal Controller for Generative Models}
%
%
\author{Enmao Diao\inst{1} \and
Jie Ding\inst{2} \and
Vahid Tarokh\inst{3}}
%
%

\institute{Duke University, Durham, NC, 27708, USA \\
\email{\{enmao.diao,vahid.tarokh\}@duke.edu} \\
\and University of Minnesota-Twin Cities, Minneapolis, MN, 55455, USA \\
\email{dingj@umn.edu}}

\maketitle              
\begin{abstract}
Class-conditional generative models are crucial tools for data generation from user-specified class labels. Existing approaches for class-conditional generative models require nontrivial modifications of backbone generative architectures to model conditional information fed into the model. This paper introduces a plug-and-play module named `multimodal controller' to generate multimodal data without introducing additional learning parameters. In the absence of the controllers, our model reduces to non-conditional generative models. We test the efficacy of multimodal controllers on CIFAR10, COIL100, and Omniglot benchmark datasets. We demonstrate that multimodal controlled generative models (including VAE, PixelCNN, Glow, and GAN) can generate class-conditional images of significantly better quality when compared with conditional generative models. Moreover, we show that multimodal controlled models can also create novel modalities of images.

\end{abstract}
\section{Introduction}
\label{sec:intro}
In recent years, many generative models based on neural networks have been proposed and achieved remarkable performance. The main backbones of generative models include Autoencoder, Autoregression, Normalization Flow, and Adversarial generative models. Perhaps the most well-known representatives of them are Variational Autoencoder (VAE) \cite{kingma2013auto}, PixelCNN \cite{oord2016pixel}, Glow \cite{kingma2018glow}, and Generative Adversarial Network (GAN)~\cite{goodfellow2014generative}, respectively. 
VAE learns a parametric distribution over an encoded latent space, samples from this distribution, and then constructs generations from decoded samples. PixelCNN uses autoregressive connections to factorize the joint image distribution as a product of conditionals over sub-pixels. Glow optimizes the exact log-likelihood of the data with a deterministic and invertible transformation. GAN was introduced as a generative framework where intractable probabilistic distributions are approximated through adversarial training. 

In many application scenarios, we are interested in constructing generations based on a conditional distribution. For instance, we may be interested in generating human face images conditional on some given characteristics of faces such as hair color, eye size, gender, etc. A systematic way to incorporate conditional information may enable us to control the data generating process with more flexibility. In this direction, conditional generative models including Conditional Variational Autoencoder (CVAE)~\cite{sohn2015learning}, Conditional Generative Adversarial Network (CGAN)~\cite{mirza2014conditional}, and Conditional PixelCNN (CPixelCNN) \cite{van2016conditional} have been proposed which model conditional information by learning the associated embeddings. The learned features are usually concatenated or added with non-conditional features at various network layers. Conditional Glow (CGlow) learns a class-conditional prior distribution and an optional auxiliary classifier.

In this paper, we propose a plug-and-play module named Multimodal Controller (MC) to allocate uniformly sampled subnetwork for each mode of data. Instead of introducing additional learning parameters to model conditional information, multimodal controlled generative models generate each mode of data from its corresponding unique subnetwork.
Our main contributions of this work are three-fold.
\begin{itemize}
\item 
We introduce a novel method to transform non-conditional generative models into class-conditional generative models, by simply attaching a multimodal controller at each layer. Unlike existing methods, our method does not introduce additional learning parameters, and it can be easily incorporated into existing implementations.
\item 
We empirically demonstrate that our method outperforms various well-known conditional generative models for datasets with small intra-class variation and a large number of data modalities. We achieve this advantage by allocating specialized subnetworks for corresponding data modalities. 
\item 
We show that multimodal controlled generative models can create novel data modalities by allocating un-trained subnetworks from genetic crossover or resampling of a codebook.

\end{itemize}

We experiment with CIFAR10, COIL100, and Omniglot datasets~\cite{krizhevsky2009learning,nene1996columbia,lake2015human}. We compare our method with the conditional generative models with different backbone generative models such as VAE, PixleCNN, Glow, and GAN. 

The rest of the paper is organized as follows. In Section 2, we review the related work. In Section 3, we introduce our proposed multimodal controller. In Section 4, we provide experimental results demonstrating the performance of our approach. Finally, we make our concluding remarks in Section 5.

\section{Related Work}
\label{sec:related}
We use four distinct backbone generative models including Variational Autoencoder (VAE)~\cite{kingma2013auto}, PixelCNN \cite{oord2016pixel}, Glow \cite{kingma2018glow} and Generative Adversarial Network (GAN)~\cite{goodfellow2014generative} to demonstrate general compatibility of our method. VAE is a directed generative model with probabilistic latent variables. PixelCNN is an autoregressive generative model that fully factorizes the joint probability density function of images into a product of conditional distributions over all sub-pixels. Glow is a flow-based generative model that enables exact and tractable log-likelihood and latent space inference by ensuring a bijective neural network mapping. GAN consists of a Generator ($G$) network and a Discriminator ($ D $) network that trains to find a Nash equilibrium for generating realistic-looking images. \newline \\
\textbf{Conditional generative models}$\,$ treat class-conditional information $h$ as input to the model. Typically, modeling such $h$ requires additional learning parameters and the resulting objective is the conditional distribution $p_\theta(x \mid h)$. Conditional VAE (CVAE)~\cite{sohn2015learning} and Conditional GAN (CGAN)~\cite{mirza2014conditional} concatenate trainable embeddings to both encoder (discriminator) and decoder (generator). Apart from trainable embeddings, this approach also requires additional learning parameters on the backbone generative models to incorporate the embeddings. Instead of concatenating, Conditional PixelCNN (CPixelCNN)~\cite{oord2016pixel} and Conditional Glow (CGlow)~\cite{kingma2018glow} adds trainable embeddings to the features. This method requires the size of embedding to match the channel size of features. There also exist many other ways of modeling $h$. ACGAN~\cite{odena2017conditional} introduces an auxiliary classifier with a class-conditional objective function to model $h$. The Conditional Normalization~\cite{dumoulin2016learned,de2017modulating} learns class-conditional affine parameters, which can be considered another way to incorporate embeddings. \cite{miyato2018cgans} proposed a projection discriminator to mode $h$ by measuring the cosine similarity between features and a learned embedding. A hybrid approach that combines the previous two methods was used in BigGAN~\cite{brock2018large}. Recently, StyleGAN~\cite{karras2019style} enables scale-specific synthesis by transforming style information with fully connected layers into affine parameters for conditional normalization. MSGAN~\cite{mao2019mode} models $h$ to generate a mode-seeking regularization term. STGAN~\cite{liu2019stgan} enables image attribute editing by modeling the difference between the source and target $h$. StyleGAN2 addresses the artifact problem caused by the adaptive instance normalization by demodulation operation~\cite{karras2020analyzing}. Latest variants of StyleGAN show that auxiliary classifier improves the performance of class-conditional generation~\cite{kang2021rebooting,sauer2022stylegan}. Transformer-based generative models leverage style vector similar to StyleGAN~\cite{park2022styleformer}. Score-based generative models modulate the generation process by conditioning on information not available during training~\cite{song2020score}.

We aim to introduce a novel and generic alternative method to generate data class-conditionally. Intuitively, training separate models for each mode of data can guarantee class-conditional data generation. However, we do not want the model complexity to grow with the number of data modalities. 

Recently, a few works pay attention to subnetworks. The lottery ticket hypothesis~\cite{frankle2018lottery,zhou2019deconstructing} states that there may exist a subnetwork that can match the performance of the original network when trained in isolation. PathNet~\cite{fernando2017pathnet} demonstrates successful transfer learning through migration from one subnetwork to the other. Piggyback and some other works in continual learning literature also adopt similar methods to train subnetworks for various tasks of data~\cite{mallya2018piggyback,rajasegaran2019random,wortsman2020supermasks}.  HeteroFL~\cite{diao2020heterofl} utilizes subnetworks in Federated Learning to reduce the computation and communication costs. \newline \\
\textbf{Multimodal controlled generative models} uniformly sample subnetworks from a backbone generative model and allocate a unique computational path for each data modality. In our paper, we refer data modality as the class-conditional modality of data rather than the type of data, i.e. language and image. Our method is different from the existing weight masking approach \cite{mallya2018piggyback,rajasegaran2019random,wortsman2020supermasks} because our mask is applied to the network representations of networks rather than the model parameters. Specifically, our approach is able to train data from multiple data modalities in the same batch of data simultaneously. However, existing weight masking methods can only optimize one data modality at a time. This paper empirically justifies the following claim. Uniformly sampled subnetworks through masking network representations can well-represent a substantial number of data modalities in class-conditional generative models by allocating a unique computational path for each mode of data.

\section{Multimodal Controller}
\label{sec:mc}
Suppose that there is a dataset $X$ with $C$ data modalities. Each mode of data $X_c = \{x^{i}_c\}^{N_c}_{i=1}$ consists of $N_c$ i.i.d. samples of a (continuous or discrete) random variable. Given a set of learning model parameters $\theta \in \mathbb{R}^{D}$ with size $D$, each mode of data is modeled with a random subset $\theta_c \subset \theta$.
For notational convenience, we will interchangeably use the notions of subset and subvector. In this way, the allocated parameters for each mode will represent both the inter-mode association and intra-mode variation, thanks to parameter sharing and specialization. 

Next, we discuss technical details in the specific context of neural networks. Suppose a uniformly sampled subnetwork takes input $X_c \in \mathbb{R}^{N_c \times K_c}$, where $N_c$ and $K_c$ are the batch size and the input channel size for data mode $c$. Suppose that the subnetwork is parameterized by a weight matrix $W_c \in \mathbb{R}^{D_c \times K_c}$ and bias vector $b_c \in \mathbb{R}^{D_c}$, where $D_c$ is the output channel size for data mode $c$, then we have output $y_c \in \mathbb{R}^{N_c \times D_c}$ where  
\begin{align*}
y_c = \phi(\text{BN}_c(X_c \times W_c^T + b_c))
\end{align*}
$\text{BN}_c(\cdot)$ denotes the Batch Normalization (BN)~\cite{ioffe2015batch} with affine parameters for corresponding data mode $c$ and $\phi(\cdot)$ is the activation function.

Existing methods~\cite{mallya2018piggyback} allocate subnetworks by masking out model parameters, i.e. $W_c = W \odot e_c$ where $e_c$ is a binary mask and $\odot$ indicates Hadamard product. However, the computation cost of the above formulation increases with the number of data modalities, as we need to sequentially backward the subnetwork of each data modality. 

Therefore we propose a nonparametric module named \textbf{Multimodal Controller (MC)} using a masking method similar to Dropout~\cite{srivastava2014dropout}, which allocates subnetworks by masking out network representations. We choose to uniformly sample the codebook because we want to create a large number of unique subnetworks. The number of unique subnetworks we can generate with MC is $2^D$. We do not train the codebook, because it is possible for some data modalities to have the same binary masks.

\begin{figure}[htbp]
\centering
  \vspace{-0.2cm}
 \includegraphics[width=0.8\linewidth]{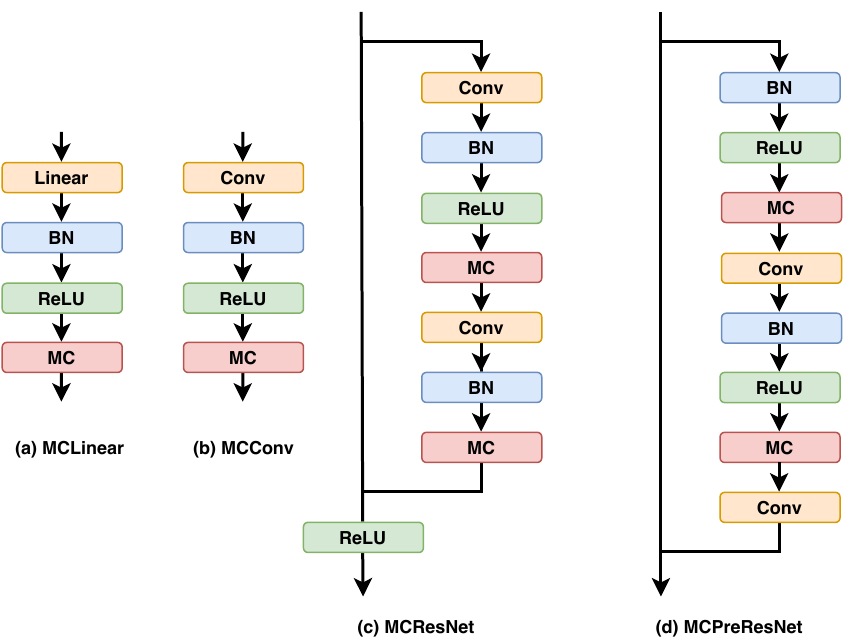}
 \caption{Multimodal Controlled Neural Networks.}
 \label{fig:fig_mc}
\end{figure}

We uniformly draw $C$ unique modality codewords $e_c \in \mathbb{F}_2^{D}$ to construct a modality codebook $e\in \mathbb{F}_2^{C \times D}$, where $\mathbb{F}_2$ denotes the binary field. Note that each row of the codebook is a binary mask allocated for each mode of data. Let $\times$ denote the usual matrix multiplication. Let $X \in \mathbb{R}^{N \times K}$ denote the output from the previous layer. Suppose that the original network is parameterized by a weight matrix $W \in \mathbb{R}^{D \times K}$ and bias vector $b \in \mathbb{R}^{D}$. Then for a specific mode of data $c$, we have multimodal controlled output $y \in \mathbb{R}^{N \times D}$ where  
\begin{align*}
\hat{W}_c &= W \odot e_c, \quad \hat{b}^{D}_c = b \odot e_c, \\
\hat{\text{BN}}_c &= \text{BN}\odot e_c, \quad  \hat{\phi}_c(\cdot) = \phi(\cdot)\odot e_c\\
\quad
y &= \hat{\phi}_c(\hat{\text{BN}}_c(X \times \hat{W}_c^T + \hat{b}_c))\\
&= \phi(\text{BN}(X \times W^T + b)) \odot e_c
\end{align*}
Note that $\hat{W}_c$, $\hat{b}^{D}_c$, $\hat{\text{BN}}_c$ and $\hat{\phi}_c$ are uniformly masked from original network. Because we are masking out the network representations, we can factorize $e_c$ to avoid interfering with the calculation of running statistics and activation function because $(e_c)^n = e_c$. Suppose we have class-conditional information $h \in \mathbb{F}_2^{N \times C}$ where each row is a one-hot vector. The multimodal controller can optimize all data modalities of one batch of data in parallel as follows
\begin{align*}
y &= \phi(\text{BN}(X \times W^T + b)) \odot (h \times e) .
\end{align*}
Suppose that the above formulation is for an intermediate layer. Then $X$ is the output masked from the multimodal controller in the previous layer. Denote $\tilde{e}_c \in \mathbb{F}_2^{K}$ as the codeword for data mode $c$ in previous multimodal controller, then the effective subnetwork weight matrix $\tilde{W}$ is $W \odot (e_c \times \tilde{e}_c)$. Since each codeword is uniformly sampled, the size of  $\tilde{W}$ is approximately $\frac{1}{4}$ the original $W$. Indeed, the multimodal controller is a nonparametric computation module modulating model architecture and trading memory space for speedup in runtime. 
Although the above formulation only describes the interaction of multimodal controller and linear network. It can be readily extended to other parametric modules, such as convolution layers. We demonstrate the suggested way of attaching this plug-and-play module in practice in Figure~\ref{fig:fig_mc}.

\section{Experiments}
\label{sec:experiments}
This section demonstrates applications of multimodal controlled generative models to data generation and creation. We compare the result of our proposed method with that of conditional generative models. We illustrate our results for four different types of multimodal controlled generative models, including VAE, PixelCNN, Glow, and GAN on CIFAR10, COIL100, and Omniglot datasets~\cite{krizhevsky2009learning,nene1996columbia,lake2015human}. Due to our proposed method's random nature, we conduct $12$ random experiments for each generative model on each dataset. The standard errors in parentheses show that uniformly sampled subnetworks are robust enough to produce stable results for either a small or large number of data modalities.

\subsection{Image Generation}
In this section, we present quantitative and qualitative results of generations from conditional and multimodal controlled generative models. More results are included in the supplementary document. Based on our results, multimodal controlled generative models can generate samples of comparable or better fidelity and diversity compared with conditional counterparts. We report our quantitative results in Table~\ref{tab:isfid} with Inception Score (IS)~\cite{salimans2016improved} and Fr\'{e}chet Inception Distance (FID)~\cite{heusel2017gans} which are perhaps the two most common metrics for comparing generative models. Both conditional and multimodal controlled generative models share the same backbone generative models for a fair comparison. \newline \\
\textbf{Quantitative Results} shown in Table~\ref{tab:isfid} demonstrate that multimodal controlled generative models performs considerably better than the conditional generative models for COIL100 and Omniglot datasets and perform comparably with conditional generative models for the CIFAR10 dataset. The major difference among them is that CIFAR10 dataset has sufficient shots for a small number of data modalities while COIL100 and Omniglot datasets only have a few shots for a large number data modalities. Note that uniformly sampled subnetworks are using approximately a quarter of the number of learning parameters as the original network. Therefore, it is difficult for a small subnetwork to learn one mode of data with a high intra-class variation. When intra-class variation is small, and the number of data modalities is large as in COIL100 and Omniglot datasets, our proposed method has a significant advantage because we can specialize subnetworks for their corresponding data modalities by allocating unique computational paths. It is worth mentioning that CPixelCNN for Omniglot outperforms MCPixelCNN because it has almost twice the learning parameters as the later. The additional parameters are used for learning conditional embeddings, which increase with the number of data modalities.

\begin{table}[htbp]
{
\centering
\caption{Inception Score (IS) and Fr\'{e}chet Inception Distance (FID) for conditional and multimodal controlled  generative models.}
\label{tab:isfid}
\resizebox{1\columnwidth}{!}{
\begin{tabular}{@{}ccccccc@{}}
\toprule
\textbf{}  & \multicolumn{2}{c}{CIFAR10}                 & \multicolumn{2}{c}{COIL100}                 & \multicolumn{2}{c}{Omniglot}                     \\ \midrule
           & IS                  & FID                   & IS                   & FID                  & IS                       & FID                   \\ \midrule
CVAE       & 3.4 (0.07)          & 133.7 (3.13)          & 89.4 (1.94)          & 37.6 (2.30)          & 539.8 (21.61)            & 367.5 (14.16)         \\
MCVAE      & \textbf{3.4 (0.05)} & \textbf{128.6 (1.32)} & \textbf{95.2 (0.50)} & \textbf{29.5 (1.91)} & \textbf{889.3 (16.94)}   & \textbf{328.5 (9.82)} \\ \midrule
CPixelCNN  & \textbf{5.1 (0.06)} & \textbf{70.8 (1.78)}  & 94.2 (0.68)          & 8.1 (0.51)           & \textbf{1048.5 (160.33)} & \textbf{23.4 (5.38)}  \\
MCPixelCNN & 4.8 (0.04)          & 75.2 (1.60)           & \textbf{98.1 (0.75)} & \textbf{4.7 (0.69)}  & 762.4 (86.07)            & 43.3 (6.20)           \\ \midrule
CGlow      & 4.4 (0.05)          & \textbf{63.9 (1.34)}  & 78.3 (2.25)          & \textbf{35.1 (2.67)} & 616.9 (6.71)             & \textbf{40.8 (1.28)}  \\
MCGlow     & \textbf{4.8 (0.05)} & 65.2 (1.21)           & \textbf{89.6 (1.63)} & 42.0 (5.19)          & \textbf{998.5 (29.10)}   & 47.2 (2.69)           \\ \midrule
CGAN       & \textbf{8.0 (0.10)} & \textbf{18.1 (0.75)}  & 97.9 (0.78)          & 24.4 (9.86)          & 677.3 (40.07)            & 51.0 (12.65)          \\
MCGAN      & 7.9 (0.13)          & 21.4 (0.83)           & \textbf{98.8 (0.09)} & \textbf{7.8 (0.44)}  & \textbf{1288.8 (5.38)}   & \textbf{23.9 (0.73)}  \\ \bottomrule
\end{tabular}
}
}
\end{table}

\begin{figure}[htbp]
\centering
 \includegraphics[width=1\linewidth]{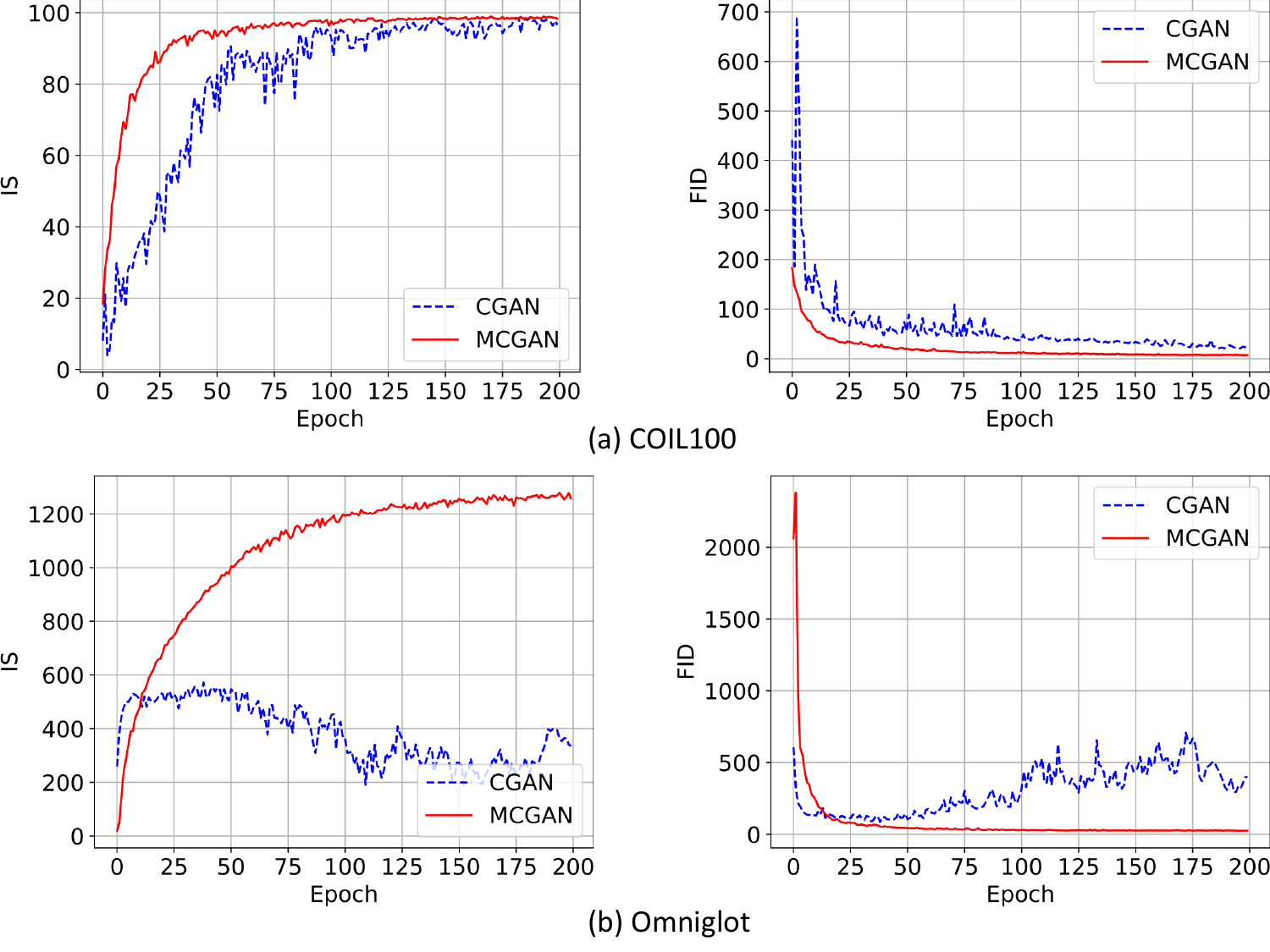}
 \caption{Learning curves of Inception Score (IS) of CGAN and MCGAN for COIL100 and Omniglot datasets.}
 \label{fig:fig_lc}
\end{figure}

\begin{figure}[htbp]
\centering
 \includegraphics[width=0.7\linewidth]{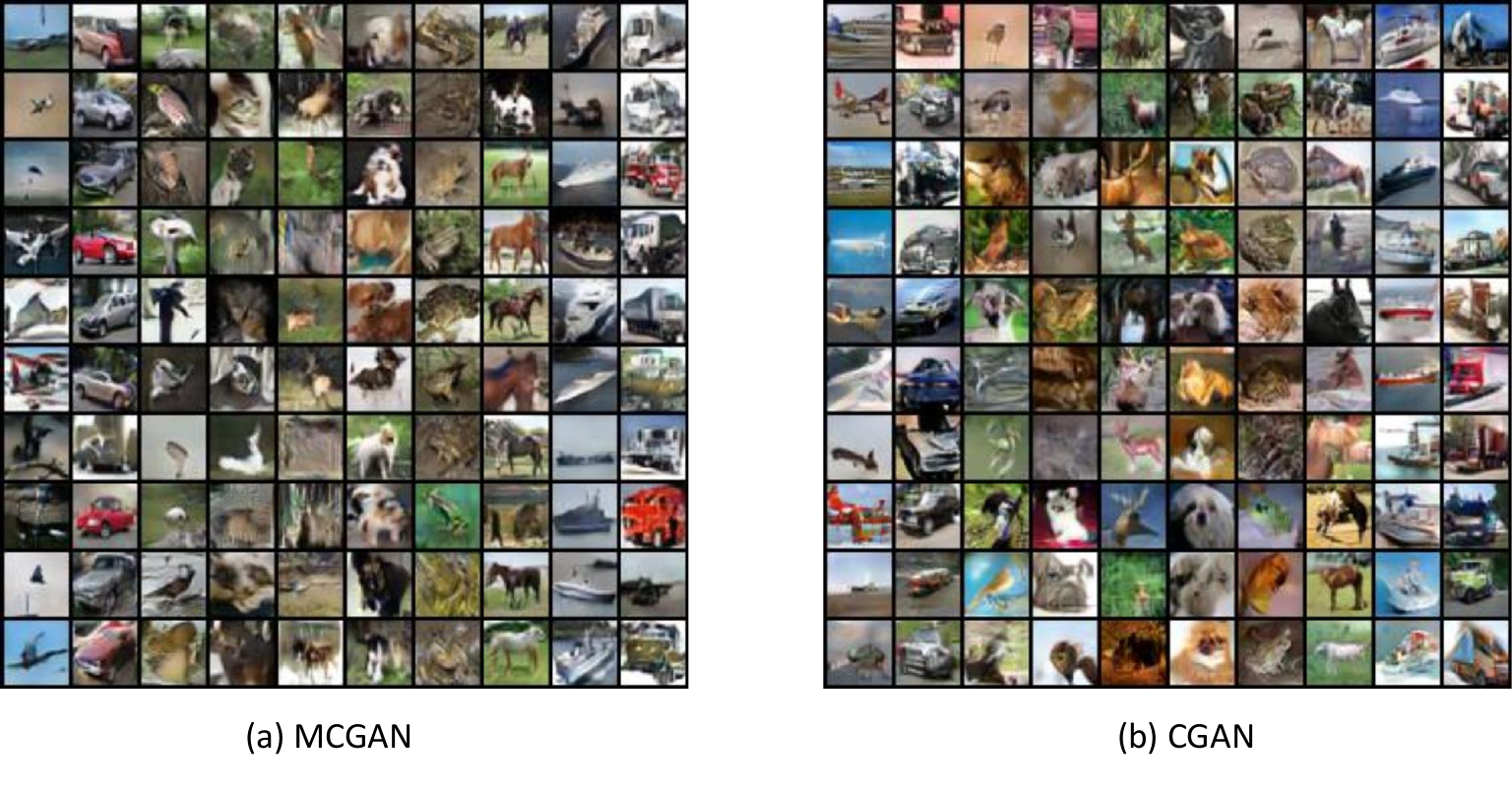}
   \vspace{-0.5cm}
 \caption{(a) MCGAN (b) CGAN trained with CIFAR10 dataset. Generations in each column are from one data modality.}
 \label{fig:Generation_GAN_CIFAR10}
\end{figure}

\begin{figure}[htbp]
\centering
 \includegraphics[width=1\linewidth]{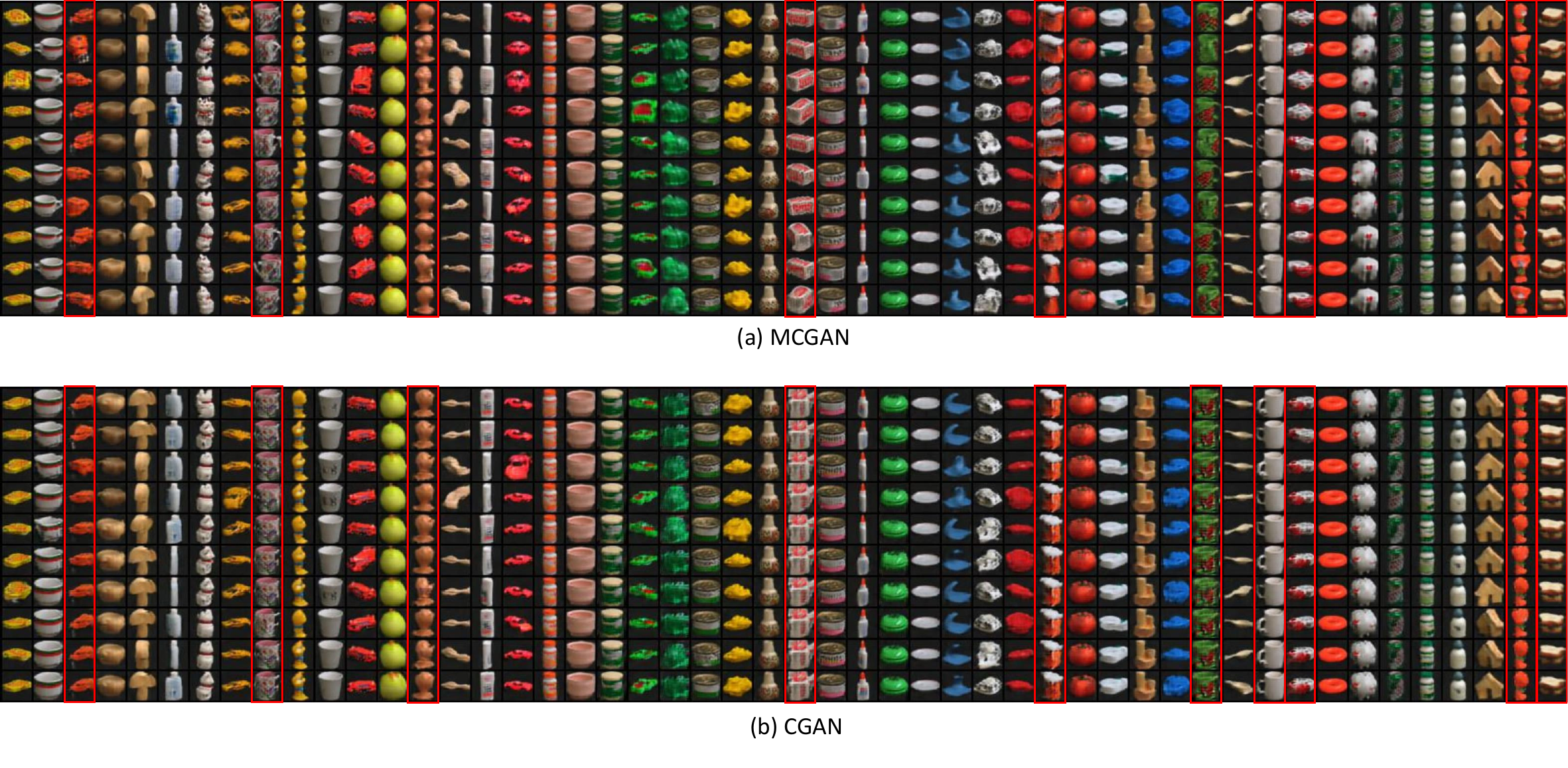}
   \vspace{-0.9cm}
 \caption{(a) MCGAN (b) CGAN trained with COIL100 dataset. Generations in each column are from one data modality.}
 \label{fig:Generation_GAN_COIL100}
\end{figure}

\begin{figure}[htbp]
\centering
 \includegraphics[width=1\linewidth]{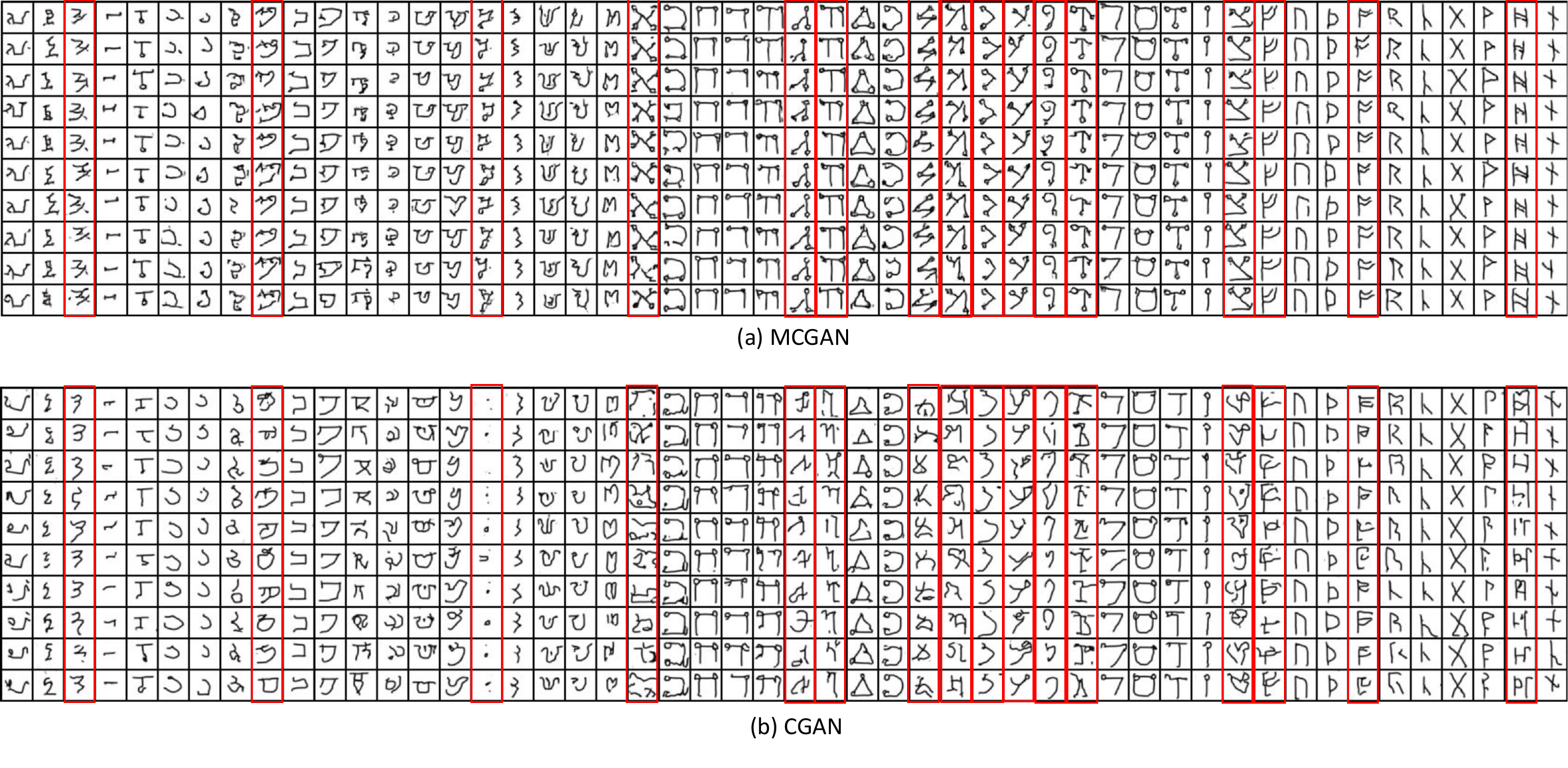}
   \vspace{-0.9cm}
 \caption{(a) MCGAN (b) CGAN trained with Omniglot dataset. Generations in each column are from one data modality.}
 \label{fig:Generation_GAN_Omniglot}
\end{figure}

We illustrate the learning curves of of MCGAN and CGAN for COIL100 and Omniglot datasets in Figure~\ref{fig:fig_lc}. The learning curves also show that MCGAN consistently outperforms CGAN. The results show that conditional generative models fail to capture and inter-class variation when the number of data modalities is large. Both our quantitative and qualitative results demonstrate the efficacy and advantage of our proposed multimodal controller for datasets with small intra-class variation and a large number of data modalities.
\newline
\textbf{Qualitative Results} regarding Multimodal Controlled GAN (MCGAN) and Conditional GAN (CGAN) for CIFAR10, COIL100, and Omniglot datasets are shown in Figure~\ref{fig:Generation_GAN_CIFAR10},~\ref{fig:Generation_GAN_COIL100}, and~\ref{fig:Generation_GAN_Omniglot} respectively. For CIFAR10 datasets, MCGAN and CGAN generate similar qualitative results. For the COIL100 dataset, the rotation of different objects can be readily recognized by MCGAN. However, CGAN fails to generate objects of diverse rotations, as shown in red lines. For the Omniglot dataset, the intra-class variation and inter-class distinction can also be identified by MCGAN. However, CGAN fails to generate some data modalities, as shown in images squared in red lines.

\subsection{Image Creation from Novel Data Modalities}
\begin{figure}[htbp]
\centering
 \includegraphics[width=0.8\linewidth]{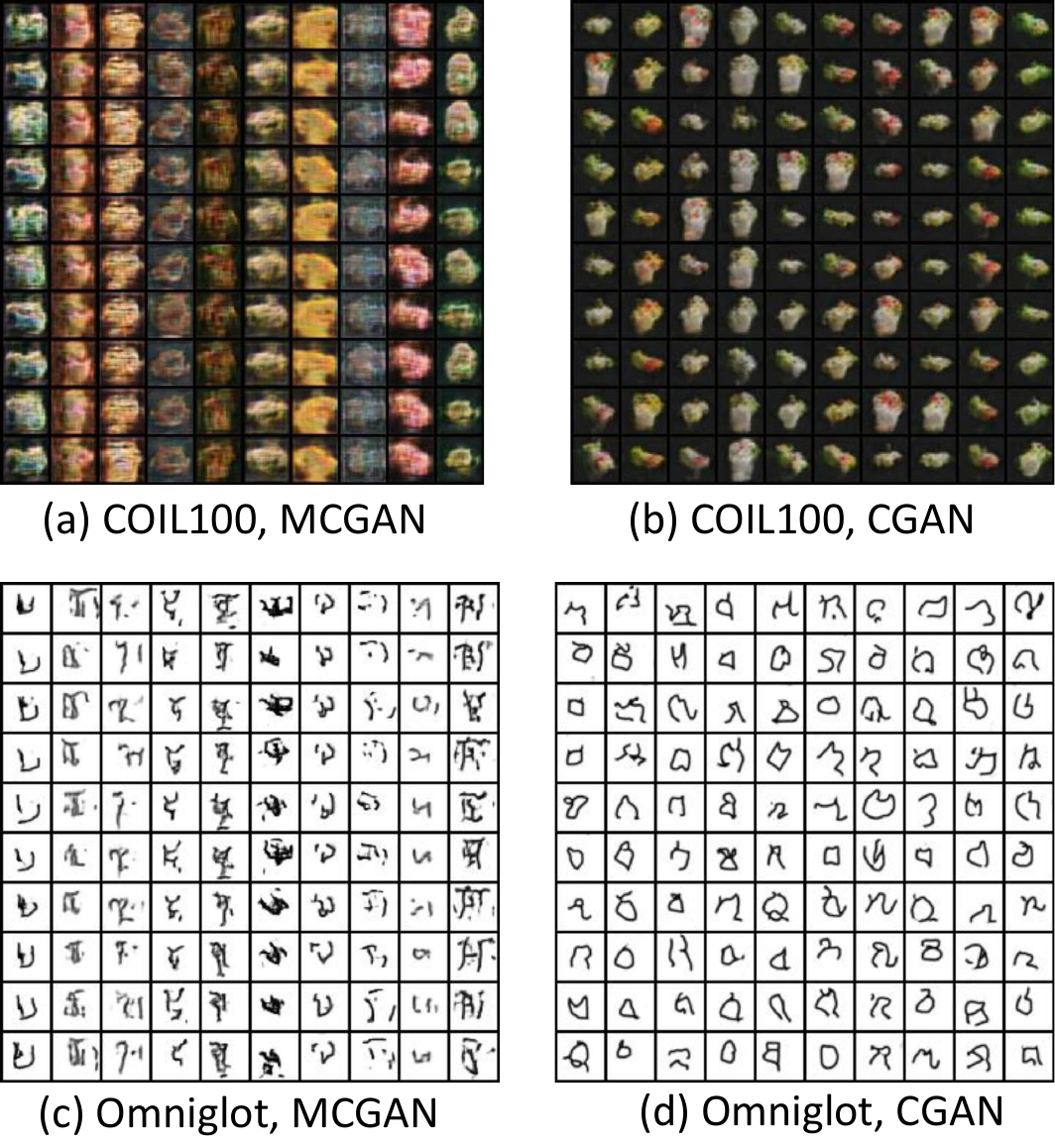}
 \vspace{-0.3cm}
 \caption{(a,b) COIL100 and (c,d) Omniglot datasets trained with MCGAN and CGAN. (a,c). We uniformly create new data modalities (each column) from pre-trained data modalities.}
 \label{fig:Creation_GAN}
\end{figure}

In this section, we provide quantitative and qualitative results of data creation from conditional and multimodal controlled generative models. We show that multimodal controlled generative models can class-conditionally synthesize from a novel data modality not prescribed in the training dataset. We propose an unbiased way of creating novel data modalities for our proposed method. Because the pre-trained codewords are uniformly sampled binary masks, we can naturally create unbiased novel data modalities by uniformly resampling the codebooks of multimodal controllers plugged in each layer. To compare with conditional generative models, we uniformly create new data modalities for conditional generative models by sampling the weights of a convex combination of pre-trained embeddings from a Dirichlet distribution.
\newline \\
\textbf{Quantitative Results} are shown in Table~\ref{tab:dbi}. We evaluate the quality of uniform data creation with Davies-Bouldin Index (DBI)~\cite{davies1979cluster}. Small DBI values indicate that data creations are closely clustered on novel data modalities. Because novel data modalities are parameterized by resampled subnetworks, data created by our proposed method can be closely clustered together. Data created by conditional generative models are not closely clustered together and have much higher DBI. It shows that a random convex combination of embeddings is not enough to create unbiased novel data modalities from pre-trained class-conditional generative models. 
\begin{table}[htbp]
\centering
\caption{Davies-Bouldin Index (DBI) for conditional and multimodal controlled generative models on the uniformly created datasets. The created dataset has the same number of modalities as the raw dataset. Small DBI values indicate that data creations are closely clustered on novel data modality.}
\label{tab:dbi}
\resizebox{0.5\columnwidth}{!}{
\begin{tabular}{@{}cccc@{}}
\toprule
            & CIFAR10             & COIL100             & Omniglot            \\ \midrule
Raw dataset & 12.0                & 2.7                 & 5.4                 \\ \midrule
CVAE        & 37.9 (5.10)         & 15.5 (0.60)         & 8.5 (0.04)          \\
MCVAE       & \textbf{2.1 (0.17)} & \textbf{1.8 (0.06)} & \textbf{3.0 (0.03)} \\ \midrule
CPixelCNN   & 27.6 (1.84)         & 17.9 (0.59)         & 9.0 (0.19)          \\
MCPixelCNN  & \textbf{3.8 (0.34)} & \textbf{4.8 (0.18)} & \textbf{4.6 (0.19)} \\ \midrule
CGlow       & 40.3 (5.09)         & 14.3 (0.87)         & 8.0 (0.02)          \\
MCGlow      & \textbf{5.4 (0.46)} & \textbf{2.8 (0.12)} & \textbf{5.1 (0.14)} \\ \midrule
CGAN        & 33.2 (3.46)         & 10.4 (2.88)         & 7.8 (0.08)          \\
MCGAN       & \textbf{2.0 (0.30)} & \textbf{1.5 (0.07)} & \textbf{3.6 (0.02)} \\ \bottomrule
\end{tabular}
}
\end{table}
\newline
\textbf{Qualitative Results} are shown in Figure~\ref{fig:Creation_GAN}. The results demonstrate that subnetworks can create unbiased novel data modalities that have never been trained before. For COIL100 and Omniglot datasets, our proposed method can create novel data modalities with high fidelity and diversity because both datasets have a large number of data modalities. In particular, the learning parameters of those resampled subnetworks have been sufficiently exploited by a large number of pre-trained subnetworks.

\section{Conclusion}
\label{sec:conclusion}
In this work, we proposed a plug-and-play nonparametric module named Multimodal Controller (MC) to equip generative models with class-conditional data generation. Unlike classical conditional generative models that introduce additional learning parameters to model class-conditional information, our method allocates a unique computation path for each data modality with a uniformly sampled subnetwork. The multimodal controller is a general method applicable to various well-known backbone generative models, and it works particularly well for a substantial number of modalities (e.g., the Omniglot challenge). Multimodal controlled generative models are also capable of creating novel data modalities. We believe that this work will shed light on the use of subnetworks for large-scale and multimodal deep learning. 


\vspace{-0.1in}
\section*{Acknowledgement}
\vspace{-0.1in}
This work was supported by the Office of Naval Research (ONR) under grant number N00014-18-1-2244, and the Army Research Office (ARO) under grant number W911NF-20-1-0222.

%
%
%
\bibliographystyle{splncs04}
\vspace{-0.1in}
\bibliography{References}

\end{document}